\pdfoutput=1

\documentclass[a4paper,conference]{IEEEtran}

\usepackage{times}
\usepackage{epsfig}
\usepackage{graphicx}
\usepackage{amssymb}
\usepackage{amsmath,amsfonts,bm}
\usepackage{multirow}
\usepackage{subfigure}
\usepackage{booktabs}
\usepackage{algorithm, algorithmic}
\usepackage[flushleft]{threeparttable}
\usepackage[numbers, sort, comma, square]{natbib}

\def\bs{\bm}

\ifCLASSINFOpdf
\else
\fi

\begin{document}
%
\title{Learning with Multiplicative Perturbations}

\author{\IEEEauthorblockN{Xiulong Yang, and Shihao Ji}
\IEEEauthorblockA{Department of Computer Science\\
Georgia State University, USA\\
xyang22@gsu.edu; sji@gsu.edu}
}


%


\maketitle

%
\IEEEpeerreviewmaketitle


\begin{abstract}
Adversarial Training (AT) and Virtual Adversarial Training (VAT) are the regularization techniques that train Deep Neural Networks (DNNs) with adversarial examples generated by adding small but worst-case perturbations to input examples. In this paper, we propose xAT and xVAT, new adversarial training algorithms that generate \textbf{multiplicative} perturbations to input examples for robust training of DNNs. Such perturbations are much more perceptible and interpretable than their \textbf{additive} counterparts exploited by AT and VAT. Furthermore, the multiplicative perturbations can be generated transductively or inductively, while the standard AT and VAT only support a transductive implementation. We conduct a series of experiments that analyze the behavior of the multiplicative perturbations and demonstrate that xAT and xVAT match or outperform state-of-the-art classification accuracies across multiple established benchmarks while being about 30\% faster than their additive counterparts. Our source code can be found at \url{https://github.com/sndnyang/xvat}
\end{abstract}

\section{Introduction}

Over the past few years, Deep Neural Networks (DNNs) have achieved state-of-the-art performance on a wide range of learning tasks. However, the success of DNNs has a high reliance on large sets of labeled examples; when trained on small datasets, DNNs plague to overfitting (if not regularized properly). For many practical applications, collecting a large amount of labeled examples is very expensive and/or time-consuming. To address this issue, researchers have investigated a host of techniques, such as Dropout~\cite{dropout14}, AT~\cite{propertiesNN14,advexample15}, VAT~\cite{vat16}, and Mixup~\cite{mixup18}, to regularize the training of DNNs. Such techniques usually augment the loss function of DNNs with a regularization term to prevent the model from overfitting when the labeled train set is small.

Several studies have found that the performance of DNNs can be improved significantly by enforcing the prediction consistency of DNNs in response to original inputs and their perturbated versions. For instance, Szegedy et al.~\cite{propertiesNN14} have found that very tiny perturbations to input samples (a.k.a., adversarial examples) can easily fool a well-trained DNN because the decision boundary of the DNN can change sharply around some data points. To improve the robustness of DNNs, they introduce AT to regularize the training of DNNs by augmenting the training set with adversarial examples. Furthermore, VAT~\cite{vat16,vat18} extends the adversarial training principle of AT from supervised learning to semi-supervised learning by generating adversarial perturbations on unlabeled examples based on a divergence measure. However, the perturbations exploited by AT and VAT are \emph{additive} in the sense that these perturbations are added pixel-wise to input examples.

\begin{figure*}[t]
    \centering
    \subfigure[\emph{Additive} Perturbation Pipeline]{
        \includegraphics[width=0.9\columnwidth]{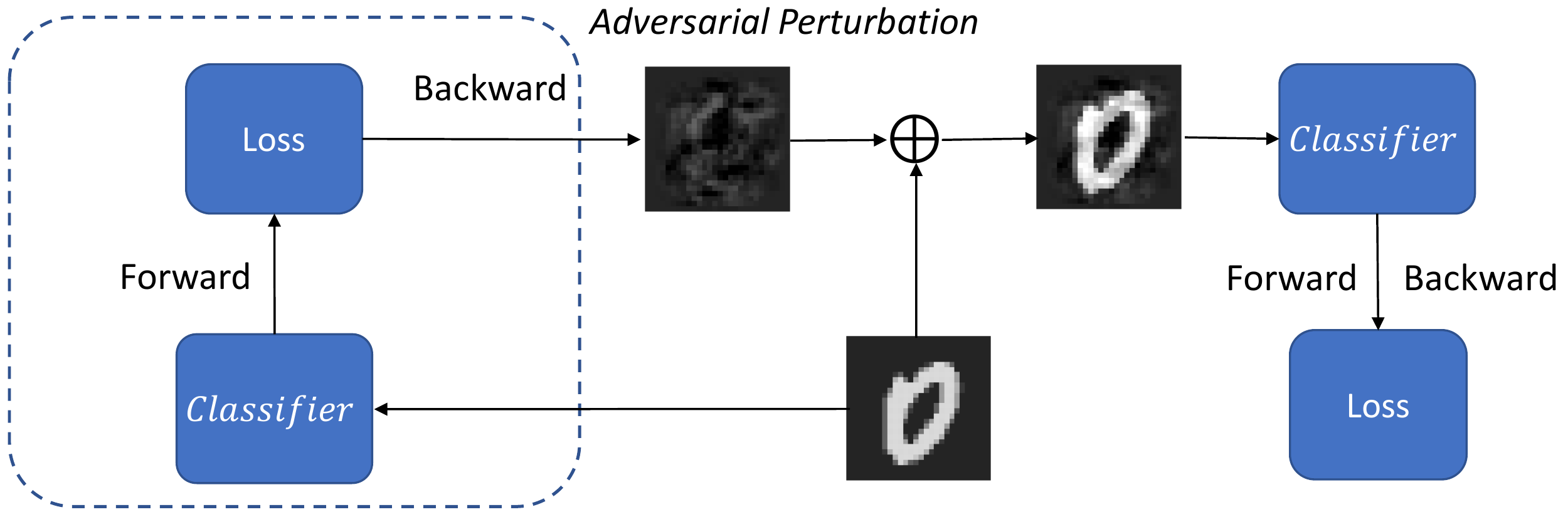}
        \label{fig:add_process}
    }
    \subfigure[\emph{Multiplicative} Perturbation Pipeline]{
        \includegraphics[width=0.9\columnwidth]{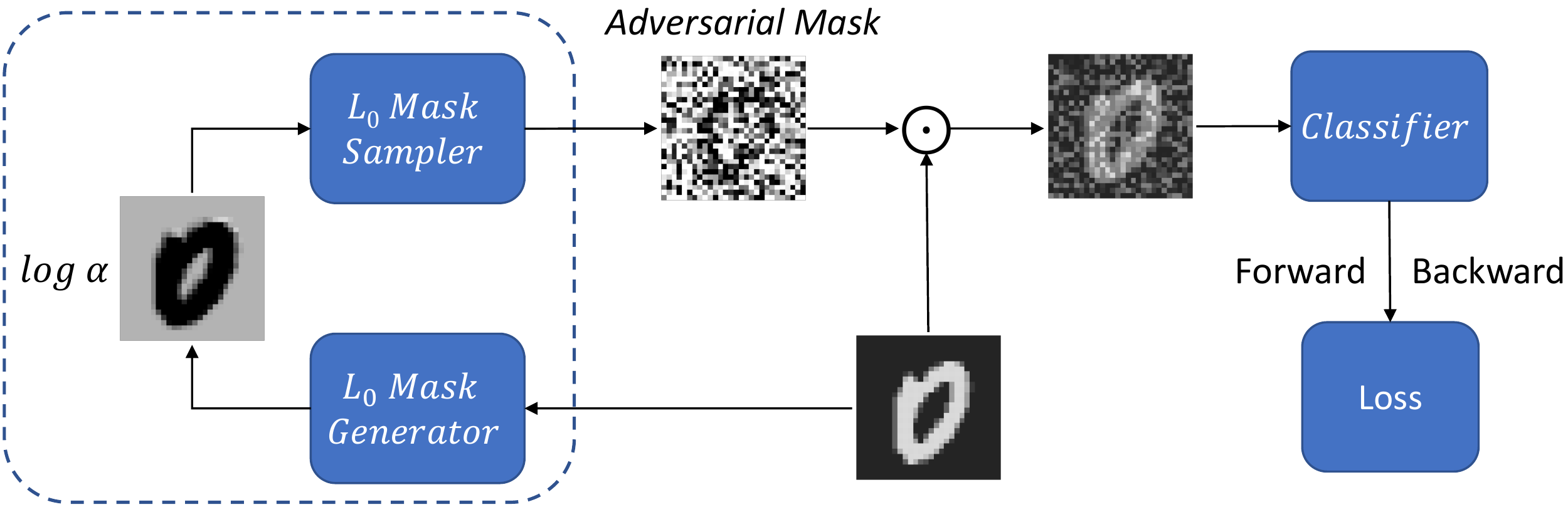}
        \label{fig:multi_process}
    }
    \vspace{-5pt}
    \caption{Comparison of the \emph{additive} perturbation pipeline and the \emph{multiplicative} perturbation pipeline.}
    \label{figure:add_multi_process}
    \vspace{-10pt}
\end{figure*}

In this paper, we propose a new type of perturbations called \emph{multiplicative} perturbations that are generated via an $L_0$-norm regularized optimization and are \emph{multiplied} to input examples pixel by pixel. This is in a stark contrast to the additive perturbations exploited by AT and VAT as the additive perturbations are generated by maximizing a divergence measure and \emph{added} to input examples pixel by pixel. To illustrate the differences, Fig.~\ref{figure:add_multi_process} demonstrates the learning pipelines of the additive perturbations and our multiplicative perturbations, with the main differences highlighted in the dashed boxes, where a pair of forward and backward propagations in the additive pipeline is replaced by a sparse mask generator in the multiplicative pipeline. Given an input image, the sparse mask generator outputs an adversarial mask, which is subsequently multiplied to the original input to generate a multiplicative adversarial example. We optimize the sparse mask generator \emph{adversarially} to maximize a divergence measure under an $L_0$-norm regularization. Similar to the additive perturbations, the multiplicative perturbations can be generated for labeled examples and unlabeled examples, and therefore can be used for supervised learning and semi-supervised learning. In light of the similarity to AT and VAT, we call our multiplicative AT and VAT as xAT and xVAT, with x denoting multiplication. Furthermore, our method can generate multiplicative perturbations transductively or inductively, while the additive perturbations are optimized transductively through backpropagation. For the reason to be discussed later, the parameters of sparse mask generator and the classifier of xAT and xVAT can be optimized simultaneously in one step, while AT and VAT have to optimize the additive perturbations and the classifier alternatively in two steps. As a result, xAT and xVAT are computationally more efficient than their additive counterparts. 

Our main contributions are summarized as follows:
\vspace{-5pt}
\begin{enumerate}
\item We introduce a new type of perturbations for robust training of DNNs that are  multiplicative instead of additive; compared to the conventional additive perturbations, the multiplicative ones are much more perceptible and interpretable;
\item The multiplicative perturbations can be generated transductively or inductively, and the sparse mask generator and the classifier can be optimized simultaneously in one step, making xAT and xVAT computationally more efficient than their additive counterparts;
\item On four image classification benchmarks, xAT and xVAT match or outperform the state-of-the-art algorithms while being about 30\% faster.
\end{enumerate}

\section{Method}
Assume that we have a labeled dataset $\mathcal{D}_l = \{(\bs{x}^i_l, y^i_l), \; i=1,2,\cdots, N_l\}$, and an unlabeled dataset $\mathcal{D}_{u} = \{\bs{x}^j_{u}, \; j=1,2,\cdots, N_{u}\}$, where $\bs{x}^k \in \mathbb{R}^P$ denotes the $k$-th input sample and $y^k$ is its corresponding label. We use $p(y|\bs{x}, \bs{\theta})$ to denote the output distribution of a classifier, parameterized by $\bs{\theta}$, in response to an input $\bs{x}$. In supervised learning, we optimize model parameter $\bs{\theta}$ by minimizing an empirical risk on $D_l$, while in semi-supervised learning both $D_l$ and $D_u$ are utilized to optimize model parameter $\bs{\theta}$.

\subsection{Additive Perturbations with AT and VAT}
AT~\cite{advexample15} and VAT~\cite{vat16} are two regularization techniques that have been proposed to generate small but worst-case perturbations for robust training of DNNs. Specifically, AT~\cite{advexample15} solves the following constrained optimization problem:
\begin{align}\label{equation:at}
  &\mathcal{L}_{\mathrm{AT}}\!\left(\bs{x}_{l}, y_{l}, \bs{r}_\text{adv}, \bs{\theta}\right)\! =\! D\left[h\!\left(y_{l}|\bs{x}_{l}\right)\!, p\!\left(y | \bs{x}_{l}+\bs{r}_{\mathrm{adv}}, \bs{\theta}\right)\right] \\
&\text{with } \bs{r}_{\mathrm{adv}} = \underset{\bs{r} ;\|\bs{r}\| \leq \epsilon}{\arg \max } D \left[h\left(y_l|\bs{x}_l\right)\!, p\left(y | \bs{x}_{l}+\bs{r}, \bs{\theta}\right)\right]\nonumber,
\end{align}
where $D[p, q]$ is a divergence measure between two distributions $p$ and $q$. For the task of image classification, $p$ and $q$ are the probability vectors whose $i$-th element denotes the probability of an input image belonging to class $i$. In AT, $D$ is the cross entropy loss $D[p, q] = -\sum_i p_i \log q_i$ and $h(y|\bs{x})$ is the one-hot encoding of label $y$ for sample $\bs{x}$. Since $h(y|\bs{x})$ requires the true label $y$ of $\bs{x}$, AT can only be applied to supervised learning. To extend the adversarial training to unlabeled samples, VAT~\cite{vat16} substitutes $h(y|\bs{x})$ with the predicted classification probability $p(y|\bs{x},\bs{\theta})$ and solves a slightly different constrained optimization problem:
\begin{align}\label{equation:vat}
  &\mathcal{L}_{\mathrm{VAT}}(\bs{x}_*, \bs{r}_\text{adv}, \bs{\theta}) \!\!=\!\! D\left[p(y | \bs{x}_{*}, \bs{\theta}), p(y | \bs{x}_{*}\!+\!\bs{r}_{\mathrm{adv}}, \bs{\theta})\right] \\
  &\text{with } \bs{r}_{\mathrm{adv}} =\underset{\bs{r} ;\|\bs{r}\|_2 \leq \epsilon}{\arg \max } D\left[p(y | \bs{x}_{*}, \bs{\theta}), p(y | \bs{x}_{*}+\bs{r}, \bs{\theta})\right]\nonumber,
\end{align}
where $\bs{x}_*$ can be either labeled data $\bs{x}_l$ or unlabeled data $\bs{x}_{u}$, and $D[p, q]$ is the KL divergence $D[p, q] = \sum_i p_i \log \frac{p_i}{q_i}$. Since no true label is required in the optimization above, VAT can be applied to both labeled and unlabeled data.

Fig.~\ref{fig:add_process} illustrates the general training pipeline of AT and VAT. Due to the constrained optimization in Eqs.~\ref{equation:at} and~\ref{equation:vat}, the exact closed-form solution of the adversarial perturbation $\bs{r}_{\mathrm{adv}}$ is intractable. Instead, fast approximation algorithms are proposed to estimate $\bs{r}_{\mathrm{adv}}$ iteratively. For AT, the adversarial perturbations can be approximated as:
\begin{equation}
    \bs{r}_{\mathrm{adv}} \approx \left\{ \begin{array}{ll}
    \epsilon \frac {\bs{g}} { \| \bs{g} \| _ { 2 }}  & {L_2\text{-norm}} \\
    \epsilon \text{sign}(\bs{g}) & {L_{\infty}\text{-norm}}, \end{array}\right.
\end{equation}
where $\bs{g} = \nabla_{\bs{x}_{l}} D \left[h(y|\bs{x}_l), p \left(y|\bs{x}_{l}, \bs{\theta} \right)\right]$. And for VAT, the perturbation can approximated via the power iteration and estimated by:
\begin{equation}
  \bs{r}_{\mathrm{adv}} \approx \epsilon \frac{\bs{g}}{\|\bs{g}\|_{2}}\quad L_2\text{-norm},
\end{equation}
where $\bs{g} = \nabla_{\bs{r}} D \left[p(y|\bs{x}_*, \bs{\theta}), p \left(y|\bs{x}_*+\bs{r}, \bs{\theta} \right)\right]$. For both algorithms, the backpropagation is needed to compute the additive perturbation $\bs{r}_\mathrm{adv}$.

\subsection{Multiplicative Perturbations}

In contrast to the additive perturbations exploited by AT and VAT, we introduce a new type of perturbations $\bs{z}$ that are multiplicative:
\begin{equation}
\bs{x}_\text{xadv}=\bs{x} \odot \bs{z},
\end{equation}
where $\bs{x}\in\mathbb{R}^P$ denotes an input image, $\bs{z} \in\{0,1\}^{P}$ is a set of binary masks, and $\odot$ is the element-wise multiplication. We can interpret $\bs{x} \odot \bs{z}$ as an operation that attaches a binary random variable $z^j$ to each pixel $j$ of $\bs{x}$, for all $ j \in \{1, 2, \cdots, P\}$. When $z^j=0$, the corresponding pixel value is set to 0. Otherwise, the corresponding pixel value is retained without any changes. With the multiplicative perturbations, we have the following constrained optimization problem:
\begin{subequations}
    \begin{equation}
\mathcal{L}_\text{xadv}(\bs{x}, \bs{z}_\text{xadv}, \bs{\theta}) = D\left[p(y | \bs{x},\bs{\theta}), p(y | \bs{x} \odot \bs{z}_\text{xadv}, \bs{\theta})\right]
\label{equation:mul_lds}
    \end{equation}
    \begin{equation}
\text{with }\bs{z}_\text{xadv} = \underset{\bs{z}}{\arg\max} D\left[p(y | \bs{x}, \bs{\theta}), p(y | \bs{x} \odot \bs{z}, \bs{\theta})\right],
\label{equation:z_1}
    \end{equation}
\end{subequations}
where $D[p, q]$ adopts the cross entropy function for xAT, and the KL divergence for xVAT. To simplify the notation, we define
\begin{equation}
  \Delta D(\bs{z}, \bs{x}, \bs{\theta}) = D\left[p(y | \bs{x},  \bs{\theta}), p(y | \bs{x} \odot \bs{z}, \bs{\theta})\right].
\end{equation}

In this formulation, $\bs{z}_\text{xadv}\in\{0,1\}^P$ is optimized adversarially to maximize the divergence measure in Eq.~\ref{equation:z_1}. This means that we wish $z^j$ to take value of 0 if zeroing out the corresponding pixel $j$ will make the prediction significantly different from the original prediction $p(y|\bs{x},\bs{\theta})$. Apparently, a trivial solution of $\bs{z}$ is all 0s, which is likely to maximize the divergence measure in Eq.~\ref{equation:z_1}, but is catastrophic to the optimization of model parameter $\bs{\theta}$ in Eq.~\ref{equation:mul_lds}. To avoid this detrimental solution, we augment Eq.~\ref{equation:z_1} with the $L_0$-norm of $\bs{z}$ to regularize the learning of $\bs{z}$:
\begin{align}
    \bs{z}_\text{xadv} &= \underset{\bs{z}}{\arg\max}\;\Delta D(\bs{z}, \bs{x}, \bs{\theta}) + \lambda\|\bs{z}\|_0  \nonumber\\
        &=\underset{\bs{z}}{\arg\max}\;\Delta D(\bs{z}, \bs{x}, \bs{\theta}) + \lambda\sum^P_{j=1} 1_{\left[z^j \neq 0 \right]}
    \label{equation:z_2}
\end{align}
where $1_{\left[c\right]}$ is an indicator function that outputs 1 if the condition $c$ is satisfied, and 0 otherwise. Consider two extreme cases. When $\bs{z}=\bs{0}$, the first term $\Delta D$ is likely to reach its maximum, but the second term $\|\bs{z}\|_0$ is minimized to 0, and thus $\bs{z}=\bs{0}$ is unlikely to maximize Eq.~\ref{equation:z_2}. On the other hand, when $\bs{z}=\bs{1}$, the first term $\Delta D$ is minimized to 0 and the second term $\|\bs{z}\|$ reaches its maximum, and thus $\bs{z}=\bs{1}$ is unlikely to maximize Eq.~\ref{equation:z_2} either. Therefore, a good solution must lie in between these two extremes, where some elements of $\bs{z}$ are 0s and the remaining are 1s. As the training proceeds, the optimized $\bs{z}$ shall gradually become an adversarial mask that blocks salient regions of an image, leads to an unreliable prediction and therefore maximizes Eq.~\ref{equation:z_2}. Subsequently, this adversarial mask will regularize the trained DNN from Eq.~\ref{equation:mul_lds} to be robust to this multiplicative perturbation. We will demonstrate this behavior when we present results.

\subsubsection{Stochastic Variational Optimization}

To optimize Eq.~\ref{equation:z_2}, we need to compute its gradient w.r.t. $\bs{z}$.
Since $\bs{z}\in\{0,1\}^P$ is a set of binary random variables, both the first term and the second term of Eq.~\ref{equation:z_2} are not differentiable. Hence, we resort to approximation algorithms to solve this binary optimization problem. We can use an inequality from stochastic variational optimization~\cite{SVO18} to derive a lower-bound of Eq.~\ref{equation:z_2}. That is, given any function $\mathcal{F}(\bs{z})$ and any distribution $q(\bs{z})$, the following inequality holds:
\begin{equation}
    \max_{\bs{z}} \mathcal{F}(\bs{z}) \geq \mathbb{E}_{\bs{z} \sim q(\bs{z})} \left[\mathcal{F}(\bs{z})\right]
\end{equation}
i.e., the maximum of a function is lower bounded by the expectation of the function.

Since $z^j, j \in\{1, 2, \cdots, P\}$ is a binary random variable, we assume $z^j$ is subject to a Bernoulli distribution with parameter $\pi^j \in \left[0, 1\right]$, i.e. $q(z^j|\pi^j) = \operatorname{Ber}\left(z^j ; \pi^{j}\right)$. Thus, Eq.~\ref{equation:z_2} can be lower bounded by its expectation:
\begin{equation}
\bs{\pi}_\text{xadv} = \underset{\bs{\pi}}{\arg \max} \mathbb{E}_{q\left(\bs{z} | \bs{\pi}\right)} \bigl[ \Delta D(\bs{z}, \bs{x}, \bs{\theta}) \bigr] +\lambda\sum^P_{j=1} \pi^j
\label{equation:z_3}
\end{equation}

Now the second term of Eq.~\ref{equation:z_3} is differentiable w.r.t. the new model parameters $\bs{\pi}$. However, the first term is still problematic as the expectation over a large number of binary random variables $\bs{z}$ is intractable and so is its gradient. Therefore, further approximations are required.

\subsubsection{The Hard Concrete Gradient Estimator}

Thanksfully, this stochastic binary optimization problem has been investigated extensively in the literature. There exist a number of gradient estimators to this problem, such as REINFORCE~\cite{reinforce92}, Gumble-Softmax~\cite{gumbel-softmax17,concrete17}, REBAR~\cite{rebar17}, RELAX~\cite{relax18} and the hard concrete estimator~\cite{l0sparse18}. We resort to the hard concrete estimator to optimize Eq.~\ref{equation:z_3} since it's straightforward to implement and demonstrates superior performance in our experiments. Specifically, the hard concrete gradient estimator employs a reparameterization trick to approximate the original optimization problem~(\ref{equation:z_3}) by a close surrogate function:
\begin{align}\label{equation:log_alpha_loss}
\log\bs{\alpha}_\text{xadv} =& \underset{\log\bs{\alpha}}{\arg \max} \mathbb{E}_{\bs{u}\sim\mathcal{U}(0,1)} \bigl[ \Delta D(g(f(\log \bs{\alpha}, \bs{u})), \bs{x}, \bs{\theta}) \bigr] \nonumber \\
     &+ \lambda\sum_{j=1}^P \sigma\left(\log\alpha^j - \beta \log \frac{-\gamma}{\zeta}\right)
\end{align}
    \vspace{-6pt}
with
\begin{align}\label{equation:log_alpha_loss_with}
&f (\log \bs{\alpha}, \bs{u} )\!=\!\sigma \left((\log \bs{u}-\log (1\!-\!\bs{u} )\!+\!\log \bs{\alpha}) / \beta \right)(\zeta\!-\!\gamma)\!+\!\gamma,  \nonumber\\
&g(\cdot) =\min (1, \max(0, \cdot)), \nonumber
\end{align}
where $\mathcal{U}(0,1)$ denotes the uniform distribution in the range of $[0,1]$, $\sigma(t) = 1 / (1+\exp(-t))$ is the sigmoid function, and $\beta = 2/3$, $\gamma = -0.1$, and $\zeta = 1.1$ are the typical values of the hard concrete distribution. With this reparameterization, the surrogate function~(\ref{equation:log_alpha_loss}) is now differentiable w.r.t. its parameters $\log\bs{\alpha}\in\mathbb{R}^P$. For more details on the hard concrete gradient estimator, we refer the readers to~\cite{l0sparse18}.

Once we have an optimized $\log\bs{\alpha}_\text{xadv}$, we can sample an adversarial mask from the hard concrete distribution $q(\bs{z}|\log\bs{\alpha}_\text{xadv})$ by
\begin{equation}
   \hat{\bs{z}}_\text{xadv}=g(f(\log\bs{\alpha}_\text{xadv}, \bs{u})), \quad\quad \bs{u}\sim\mathcal{U}(0,1).
   \label{equation:z_from_alpha}
\end{equation}
We can then optimize model parameter $\bs{\theta}$ by minimizing the following regularized empirical risk over all training examples:
\begin{align}\label{equation:final_loss}
&\mathcal{L}(\bs{\theta})=\frac{1}{N_l}\sum_{(\bs{x}_l,y_l)\in\mathcal{D}_l}\mathcal{L}_\textrm{ce}(h(\bs{x}_l,\bs{\theta}), y_l) \nonumber\\
&+ \eta\frac{1}{N_l+N_u}\sum_{\bs{x}\in\{\mathcal{D}_l,\mathcal{D}_u\}}\mathcal{L}_\text{xadv}(\bs{x}, \hat{\bs{z}}_\text{xadv},\bs{\theta}),
\end{align}
where $h(\bs{x}_l, \bs{\theta})$ denotes the prediction of a classifier, parameterized by $\bs{\theta}$, $\mathcal{L}_\textrm{ce}$ is the cross entropy loss over labeled training examples, and $\eta$ is a regularization hyperparameter that balances the cross entropy loss $\mathcal{L}_\textrm{ce}$ of labeled data and the adversarial training loss $\mathcal{L}_\text{xadv}$ of all training examples.

\begin{figure*}[ht]
    \centering
    \includegraphics[width=1.5\columnwidth]{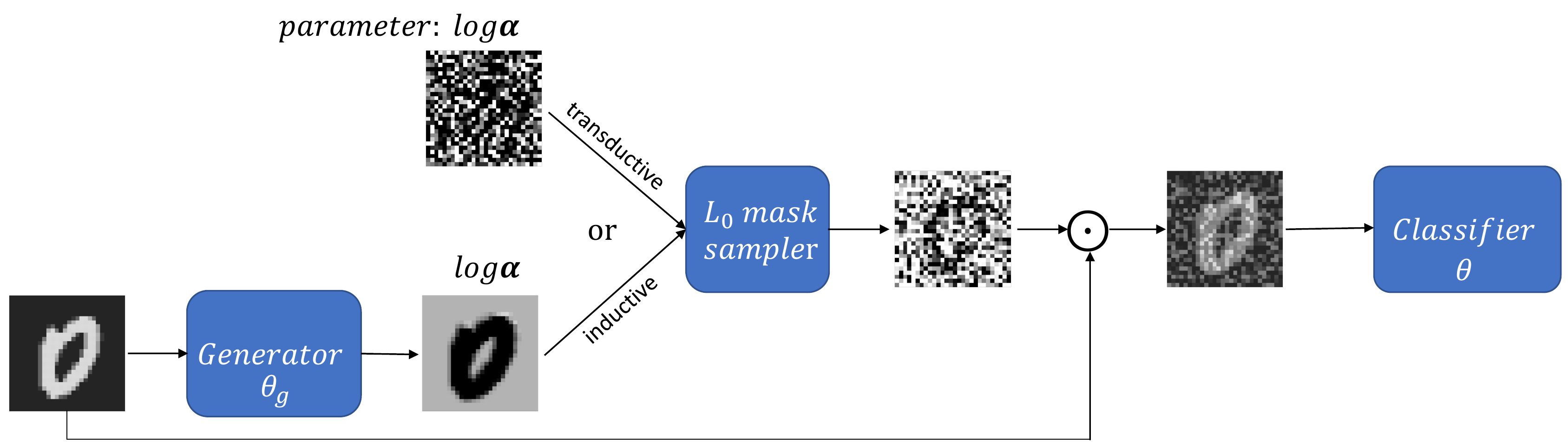}
    \caption{The pipeline of the transductive and inductive implementations of multiplicative adversarial training.}
    \label{figure:trans_ind}
    \vspace{-10pt}
\end{figure*}

\subsubsection{Transductive vs. Inductive Training}
The optimization of the final loss function~(\ref{equation:final_loss}) can be implemented transductively or inductively. For a transductive implementation, we set $\log\bs{\alpha} \in \mathbb{R}^P$ as model parameters to be optimized for each input example $\bs{x} \in \{\mathcal{D}_l, \mathcal{D}_u\}$. The learned $\log\bs{\alpha}$ is then used to generate a sparse adversarial mask $\hat{\bs{z}}_\text{xadv}$, and subsequently the training proceeds to evaluate the final loss~(\ref{equation:final_loss}). However, this approach can't generate masks for images never seen during training. A more appealing approach is the inductive implementation, which employs a generator to produce a sparse adversarial mask for any input example $\bs{x}$. To formulate the inductive training, we model the generator as a neural network, parameterized by $\bs{\theta}_g$, and given an input $\bs{x}$ we define its output as $\log\bs{\alpha} = G(\bs{x}, \bs{\theta}_{g})$, which is used subsequently to sample a sparse adversarial mask $\hat{\bs{z}}_\text{xadv}$. The pipeline of the two implementations is illustrated in Fig.~\ref{figure:trans_ind}, which can be optimized end-to-end.

One advantage of xAT/xVAT over AT/VAT is that due to the hard concrete reparameterization, we can update (i) classifier parameter $\bs{\theta}$, and (ii) the hard concrete parameter $\log\bs{\alpha}$ (for transductive training) or the generator parameter $\bs{\theta}_g$ (for inductive training) simultaneously in one step (See Fig.~\ref{figure:trans_ind}). This is not true for AT and VAT since both of them rely on backpropagation to optimize additive perturbations, while backpropagating through the perturbations produced by backpropagation is computationally unstable. Therefore, both AT and VAT resort to optimizing additive perturbations $\bs{r}_\text{adv}$ and classifier parameter $\bs{\theta}$ alternatively in two steps. Because of this advantage, xAT and xVAT are computationally more efficient than their additive counterparts. We will demonstrate this  when we present results (Table~\ref{semi-time}).

\subsubsection{Shrink or Expand Multiplicative Perturbations}

\begin{figure}[ht!]
    \vspace{-10pt}
    \centering
    \subfigure[Additive Perturbation]{
        \includegraphics[width=0.45\columnwidth]{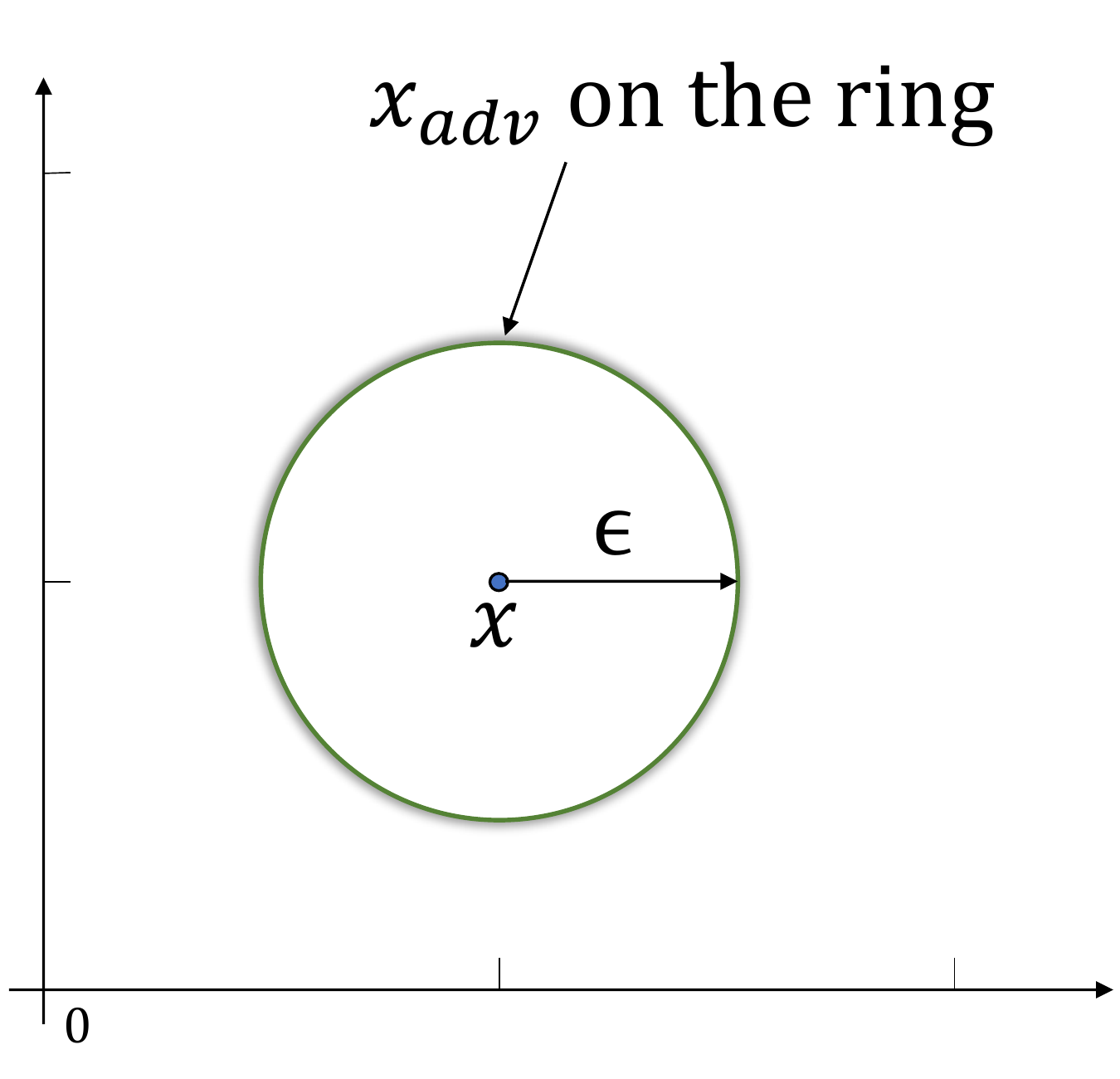}
        \label{figure:eps_of_add}
    }
    \subfigure[Multiplicative Perturbation]{
        \includegraphics[width=0.45\columnwidth]{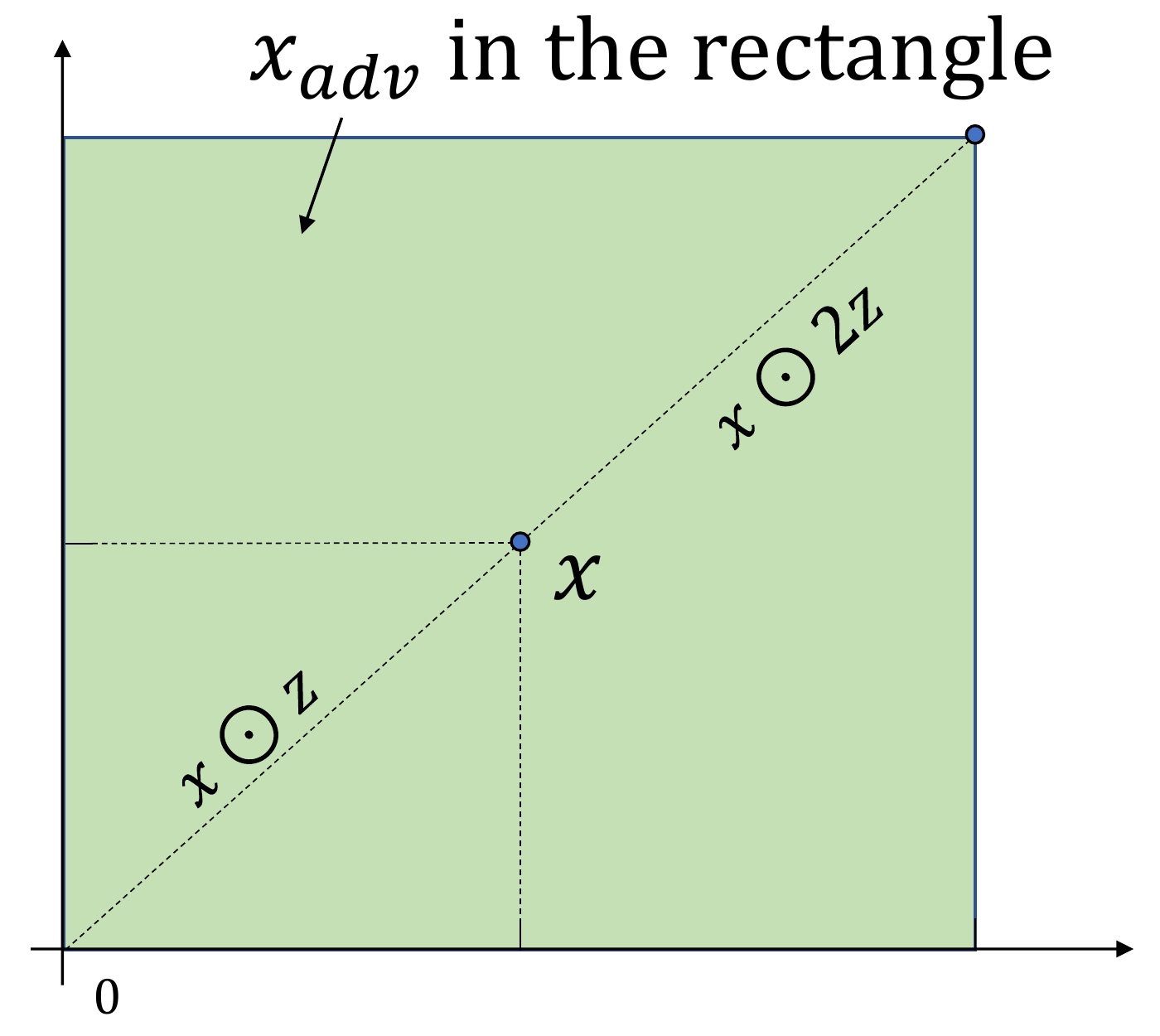}
        \label{figure:eps_of_multi}
    }
    \vspace{-5pt}
    \caption{The effect of $\epsilon$ on different perturbations. (a) shows that the additive perturbations are on the surface of a ball with the radius $\epsilon$. (b) demonstrates that our multiplicative perturbations are distributed within the rectangle.}
    \label{figure:eps_in_pert}
    \vspace{-5pt}
\end{figure}

It is worth mentioning one potential issue of multiplicative perturbations. Consider the additive adversarial example $\bs{x} + \bs{r}_{\mathrm{adv}}$ with $\bs{r}_{\mathrm{adv}} \approx \epsilon \frac{g}{\|g\|_p}$ (where $p=2$ or $\infty$), and the multiplicative adversarial example $\bs{x} \odot \bs{z}$ with $\bs{z}\in\{0,1\}^P$. Apparently, these two types of perturbations have very distinct geometric interpretations w.r.t. input example $\bs{x}$. For additive perturbations, the approximation $\bs{r}_{\mathrm{adv}} \approx \epsilon \frac{g}{\|g\|_{p}}$ indicates that all additive perturbations are generated exactly on the surface of a ball (or a box) centered at $\bs{x}$ with a radius of $\epsilon$, and no additive perturbations would appear inside the ball (see Fig.~\ref{figure:eps_of_add}). On the other hand, the multiplicative perturbations do not have this characteristic. Since $z^j$ is between 0 and 1, the admissible multiplicative perturbations can appear inside a box rather than only on the surface of a box (see Fig.~\ref{figure:eps_of_multi}). Note that the following inequality always holds:
\vspace{-1mm}
\begin{equation}
  \|\bs{x} \odot \bs{z}\| \leq \|\bs{x}\|, \text{   with    } \bs{z}\in[0, 1]^P.
\vspace{-2mm}
\end{equation}
What does this mean is that the masked image $\bs{x}\odot\bs{z}$ is always  darker than the original image $\bs{x}$, which may be undesirable in some cases where the contrast of an image is low. To tackle this potential issue, we introduce a new parameter $\epsilon$ in the multiplicative formulation: $\bs{x} \odot \epsilon\bs{z}$. We only consider $\epsilon=1$ or $2$ in our experiments, while other positive values of $\epsilon$ are admissible. When $\epsilon=1$, the masked images can only be darker than the original images, while when $\epsilon=2$ the masked images can be darker or brighter as illustrated in Fig.~\ref{figure:eps_of_multi}. Empirically, we find that $\epsilon=1$ works reasonable well on all benchmark datasets in our experiments. We therefore don't tune it any further.

Comparing Fig.~\ref{figure:eps_of_add} and Fig.~\ref{figure:eps_of_multi}, it is clear that the solution space of multiplicative perturbations is much larger than that of additive perturbations. As we will show in the experiments, due to this difference, the trained DNNs from both perturbations demonstrate very distinct weight distributions.

\begin{figure*}[t]
    \centering
    \includegraphics[width=0.8\textwidth]{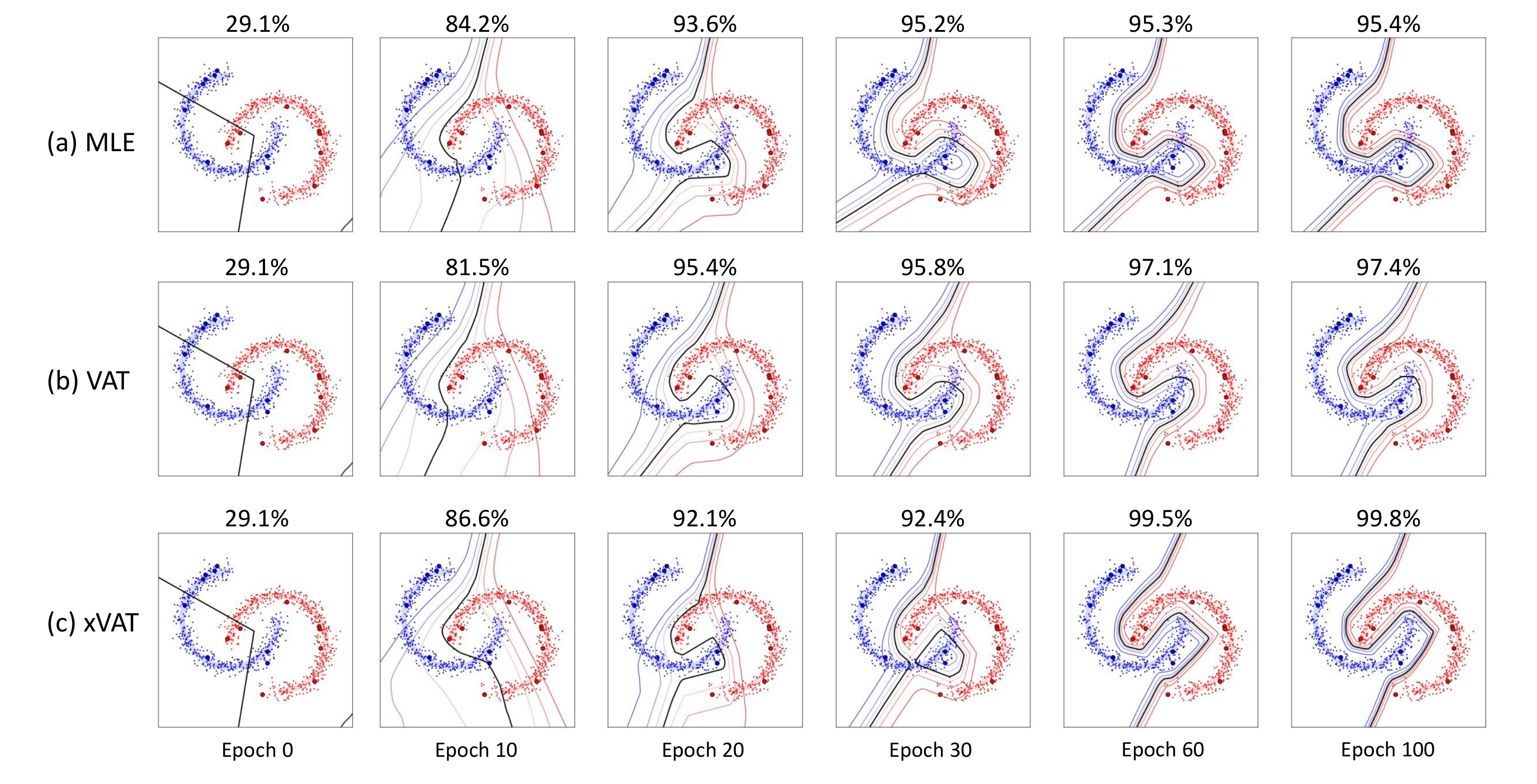}\vspace{-10pt}
    \caption{The evolution of the decision boundaries of MLE, VAT and xVAT on the synthetic ``moons" dataset.}
    \label{figure:moon_process}
    \vspace{-15pt}
\end{figure*}

\section{Related Work}

To improve the generalization of trained DNNs on unseen data examples, a variety of regularization techniques have been proposed in the past few years. Traditional regularization techniques impose an extra term to prevent the model from overfitting to a small set training examples. 
Recenlty, a family of regularization techniques called consistency regularization has been proposed that encourages classifiers to yield consistent predictions on data samples and their perturbed versions. For example, the $\Pi$-Model~\cite{temporalensembling2017} and STP~\cite{stochtranspert2016} incorporate the following consistency loss
\begin{equation}
\| p(y | \text{Augment}_1(\bs{x}), \bs{\theta}) - p(y | \text{Augment}_2(\bs{x}), \bs{\theta}) \|_{2}^{2}
\label{equation:cons_aug}
\end{equation}
as a regularization for robust training of DNNs, where $\text{Augment}(\bs{x})$ represents a stochastically perturbed version of $\bs{x}$ so that the two predictions in Eq.~\ref{equation:cons_aug} are noisy, and minimizing their difference may improve the robustness. 

Similarly, Temporal Ensembling~\cite{temporalensembling2017}, Mean Teacher~\cite{meanteachers17} and fast-SWA~\cite{swa19} generate stochastic outputs by tracking the exponential moving average (EMA) of the past predictions and weights. AT and VAT generate small but worse-case perturbations as data augmentation. All of these approaches encourage classifiers to produce consistent predictions on different perturbed examples. More recently, by combining the dropout and adversarial training, Park et al.~\cite{advdrop18} propose the adversarial dropout, which embeds an adversarial dropout layer to DNNs and  optimizes the dropout mask adversarially to maximize the loss of DNNs. This work has a similar high-level idea to ours. However, their method works on much smaller last FC layer, and employs an expensive integer programming to solve the optimization problem. Because of these limitations, their technique cannot be applied to earlier but larger layers, while our xAT and xVAT operate at input layer with an $L_0$-regularized binary optimization, which can be optimized efficiently via the hard concrete gradient estimator. Our experiments verify that xVAT not only demonstrates the state-of-the-art accuracies but also is about 30\% faster than VAT, and therefore is much more scalable.

\section{Experimental Results}
We validate our xAT/xVAT on multiple datasets with different network architectures for supervised learning and semi-supervised learning. Specifically, we illustrate how xVAT learns on a synthetic ``moons" dataset~\cite{vat16}. We also demonstrate xAT and xVAT on four established image classification benchmarks: MNIST~\cite{mnist}, SVHN~\cite{svhn11}, CIFAR-10 and CIFAR-100~\cite{cifar09}. Our main baselines are AT~\cite{advexample15} and VAT~\cite{vat18} since our work is primarily to investigate the effectiveness of multiplicative perturbations \emph{vs.} additive perturbations. We also compare xAT/xVAT with many other algorithms in terms of classification accuracy. For a fair comparison, our implementation and experimental setup~\footnote{\url{https://github.com/sndnyang/xvat/}} closely follows that of VAT~\footnote{\url{https://github.com/takerum/vat_tf/}}.

\subsection{Synthetic Dataset}
We first demonstrate the performance of xVAT on a synthetic ``Moons" dataset~\cite{vat16}. The dataset contains 16 labeled training data points and 1000 test data points, uniformly distributed in two moon-shaped clusters for binary classification. We randomly select 500 test points as unlabeled training examples and use all 1000 test points for evaluation. We compare the results of Maximum Likelihood Estimation (MLE) using the cross-entropy loss, VAT and xVAT. MLE only uses 16 labeled data points for supervised training, while VAT and xVAT use 16 labeled and 500 unlabeled data points for semi-supervised training. Fig.~\ref{figure:moon_process} illustrates the evolution of the decision boundaries of MLE, VAT, xVAT on the ``moons" dataset. As can be seen, xVAT yields a similar or slightly better decision boundary than that of VAT. Furthermore, both VAT and xVAT lean the decision boundaries that respect the underlying data manifold, demonstrating the effectiveness of multiplicative perturbations on this synthetic dataset.

\vspace{-1pt}
\subsection{Image Classification Benchmarks}
We next evaluate xAT and xVAT on four established image classification benchmarks. Detailed description of the benchmarks, network architectures and experimental setup for reproducibility can be found in supplementary material and our open source implementation.

\subsubsection{Semi-supervised Learning}
For MNIST, we use an MLP with four hidden layers of 1200, 600, 300 and 150 units, respectively, which is the same architecture used in VAT~\cite{vat16,vat18}. For SVHN, CIFAR-10 and CIFAR-100, the network architecture is a 13-layer CNN, the same as the one used in~\cite{temporalensembling2017,vat18,advdrop18}. It is worth mentioning that for xAT and xVAT we normalize the MNIST images to $[-0.5, 0.5]$ rather than $[0, 1]$. This is because multiplicative perturbations have no effect on zero-valued pixels, while the MNIST images happen to have zero-valued black background, and therefore the $[-0.5, 0.5]$ normalization allows xAT/xVAT to generate valid multiplicative perturbations on the background of MNIST. On the other datasets, we adopt the same normalization approaches as in VAT. For the transductive implementation, we initialize the parameter $\log\bs{\alpha}$ for each image with samples from a unit Gaussian distribution $\mathcal{N}(0, 1)$. For the inductive implementation, we use an one-layer CNN with one $3\times 3$ filter as the generator. Interestingly, such a simple generator is sufficient to yield competitive results for all our experiments.

\begin{table}[h!]
\caption{Test accuracies of semi-supervised learning on MNIST, SVHN and CIFAR-10. The results are averaged over 5 runs.}
\label{semi-bench}\vspace{-10pt}
\begin{center}
\setlength\tabcolsep{1.5pt}
\begin{threeparttable}
\begin{tabular}{lccc}
\toprule
\multirow{3}{*}{Method}               & \multicolumn{3}{c}{Test Accuracy (\%)}  \\
                                      & MNIST       & SVHN        & CIFAR-10    \\
                                      & $N_l$=100   & $N_l$=1000  & $N_l$=4000 \\
\midrule
GAN with feature match \cite{imprgan16}               & \textbf{99.07}       & 91.89       & 81.37       \\
CatGAN  \cite{cganulssl15}                               & 98.09       &  -          & 80.42       \\
Ladder Networks \cite{laddernetworks15}                      & 98.94       &  -          & 79.60       \\
$\Pi$-model \cite{temporalensembling2017}                           &  -          & 94.57       & 83.45       \\
Mean Teacher \cite{meanteachers17}                         &  -          & \textbf{94.79}       & 82.26       \\
VAT \cite{vat18}                                  & 98.64       & 94.23       & 85.18       \\
\midrule
xVAT (Transductive)                        & 98.02       & 93.99       & 85.82       \\
xVAT (Inductive)                       & 97.82       & 94.22       & \textbf{86.59}       \\
\bottomrule
\end{tabular}
\end{threeparttable}
\end{center}
\vspace{-10pt}
\end{table}

\begin{table}[h!]
\vspace{-5pt}
\caption{Test accuracies of semi-supervised learning algorithms on CIFAR-100. The results are averaged over 5 runs.}
\label{semi-CIFAR-100}\vspace{-5pt}
\begin{center}
\begin{threeparttable}
\begin{tabular}{lc}
\toprule
\multirow{2}{*}{Method} & Test Accuracy (\%) \\
                        & $N_l$=10000        \\
\midrule
Supervised \cite{temporalensembling2017}              & 55.44              \\
$\Pi$-model \cite{temporalensembling2017}             & 60.81              \\
Temporal ensembling \cite{temporalensembling2017}     & 61.35              \\
VAT \cite{vat18}                    & 59.71              \\
\midrule
xVAT (Transductive)     & 61.30              \\
xVAT (Inductive)        & \textbf{61.76}     \\
\bottomrule
\end{tabular}
\end{threeparttable}
\end{center}
\vspace{-15pt}
\end{table}

\begin{table}[h!]
\newcommand{\tabincell}[2]{\begin{tabular}{@{}#1@{}}#2\end{tabular}}
\caption{The training speeds of VAT and xVAT on the four benchmark datasets. The results are averaged over 5 runs.}
\label{semi-time}\vspace{-10pt}
\begin{center}
\setlength\tabcolsep{2pt}
\begin{threeparttable}
\begin{tabular}{lcccc}
\toprule
\multirow{2}{*}{Method}      & \multicolumn{4}{c}{Seconds per epoch}        \\ \\[-1em]
                             & MNIST   & SVHN    & CIFAR-10   & CIFAR-100   \\
\midrule
\tabincell{c}{VAT (ours)$^*$}                   & 4.31    & 54.3    & 51.3       & 51.5        \\
\tabincell{c}{xVAT (Transductive)}                & 4.54    & 36.6    & 34.1       & 39.3        \\
\tabincell{c}{xVAT (Inductive)}                & 4.33    & 35.7    & 33.6       & 34.4        \\
\bottomrule
\end{tabular}
\begin{tablenotes}
  \scriptsize\item $^*$For a fair comparison, we implement VAT in PyTorch and achieve similar accuracies as the official VAT implementation.
\end{tablenotes}
\end{threeparttable}
\end{center}
\vspace{-15pt}
\end{table}

Tables~\ref{semi-bench} and \ref{semi-CIFAR-100} summarize the classification accuracies of xVAT on the four benchmark datasets as compared to the state-of-the-art algorithms. As we can see, xVAT achieves very competitive and sometimes even better accuracies than the state-of-the-arts, such as the $\Pi$-model~\cite{temporalensembling2017} and VAT. Furthermore, the inductive xVAT outperforms the transductive xVAT on 3 out of 4 benchmarks. This is likely because the per-image parameter $\log\bs{\alpha}$ is more difficult to optimize than that of the shared generator parameter $\bs{\theta}_g$. Table~\ref{semi-time} compares the per-epoch training speeds of VAT and xVAT. xVAT (both transductive and Inductive) is about 30\% faster than VAT. This is because the generator of xVAT is significantly shallower than the CNN classifier, and therefore the multiplicative perturbations can be generated much more efficiently than the backpropagation-based additive perturbations. In general, VAT requires at least two pairs of forward and backward propagations per iteration, while xVAT only needs one pair of forward and backward propagation, plus some extra time to update the parameter of generator. In our experiments, since an one-layer CNN generator is sufficient, which is significantly cheaper than the 13-layer CNN classifier, the cost of updating the generator is largely negligible, and therefore our xVAT is computationally more efficient than VAT. Because inductive xAT/xVAT in general outperform their transductive ones, in the experiments that follow we use inductive xAT/xVAT unless noted otherwise.

\begin{table}[h!]
\caption{Test accuracies of supervised learning on MNIST. The results are averaged over 5 runs.}
\label{sup-MNIST}\vspace{-10pt}
\begin{center}
\begin{threeparttable}
\begin{tabular}{lcc}
\toprule
Method                                     & Test Accuracy (\%) \\
\midrule
Dropout~\cite{dropout14}                   & 98.95 \\
Concrete Dropout~\cite{concretedrop17}     & 98.60 \\
Ladder networks~\cite{laddernetworks15}    & \textbf{99.43} \\
\midrule
Baseline (MLE) \cite{vat18}                            & 98.89     \\
AT, $L_{\infty}$  \cite{vat18}                & 99.21     \\
AT, $L_{2}$  \cite{vat18}               & 99.29     \\
VAT~\cite{vat18}                     & 99.36     \\
\midrule
xAT (Inductive)                         & 99.18     \\
xVAT (Inductive)                        & 99.08     \\
\bottomrule
\end{tabular}
\end{threeparttable}
\end{center}
\vspace{-10pt}
\end{table}

\begin{table}[h!]
\vspace{-2pt}
\caption{Test accuracies of supervised learning on CIFAR-10 and CIFAR-100. The results are averaged over 5 runs.}
\label{sup-CIFAR-10}\vspace{-5pt}
\begin{center}
\begin{threeparttable}
\begin{tabular}{lcc}
\toprule
\multirow{2}{*}{Method}                 & \multicolumn{2}{c}{Test Accuracy (\%)} \\
                                        & CIFAR-10   & CIFAR-100    \\
\midrule
Baseline (MLE) \cite{temporalensembling2017}                         & 93.24     & 73.58        \\
$\Pi$-model   \cite{temporalensembling2017}                           & \textbf{94.44}     & 73.68        \\
Temporal ensembling \cite{temporalensembling2017}                    & 94.40   & 73.70        \\
AT, $L_{\infty}$(ours)*               & 93.90     &  74.04    \\
VAT \cite{vat18}                             & 94.19     & 75.02        \\
\midrule
xAT (Inductive)                         & 93.70     & 74.62        \\
xVAT (Inductive)                        & 93.88     & \textbf{75.30}        \\
\bottomrule
\end{tabular}
\begin{tablenotes}
  \scriptsize\item $^*$No reported results. For a fair comparison, we implement AT in PyTorch and verify its accuracies on MNIST, and report our results in the table.
\end{tablenotes}
\end{threeparttable}
\end{center}
\vspace{-10pt}
\end{table}

\subsubsection{Supervised Learning}
We next evaluate the performance of xAT and xVAT in supervised learning on MNIST, CIFAR-10 and CIFAR-100. The same network architectures are used as in the experiments of semi-supervised learning. We compare the results of xAT and xVAT with the state-of-the-art algorithms in Tables~\ref{sup-MNIST} and~\ref{sup-CIFAR-10}. It is demonstrated that our xAT and xVAT achieve very competitive accuracies to the other competing methods, demonstrating the effectiveness of multiplicative perturbations in supervised learning.

\begin{figure*}[t]
    \centering
    \subfigure[]{
        \includegraphics[width=0.88\columnwidth]{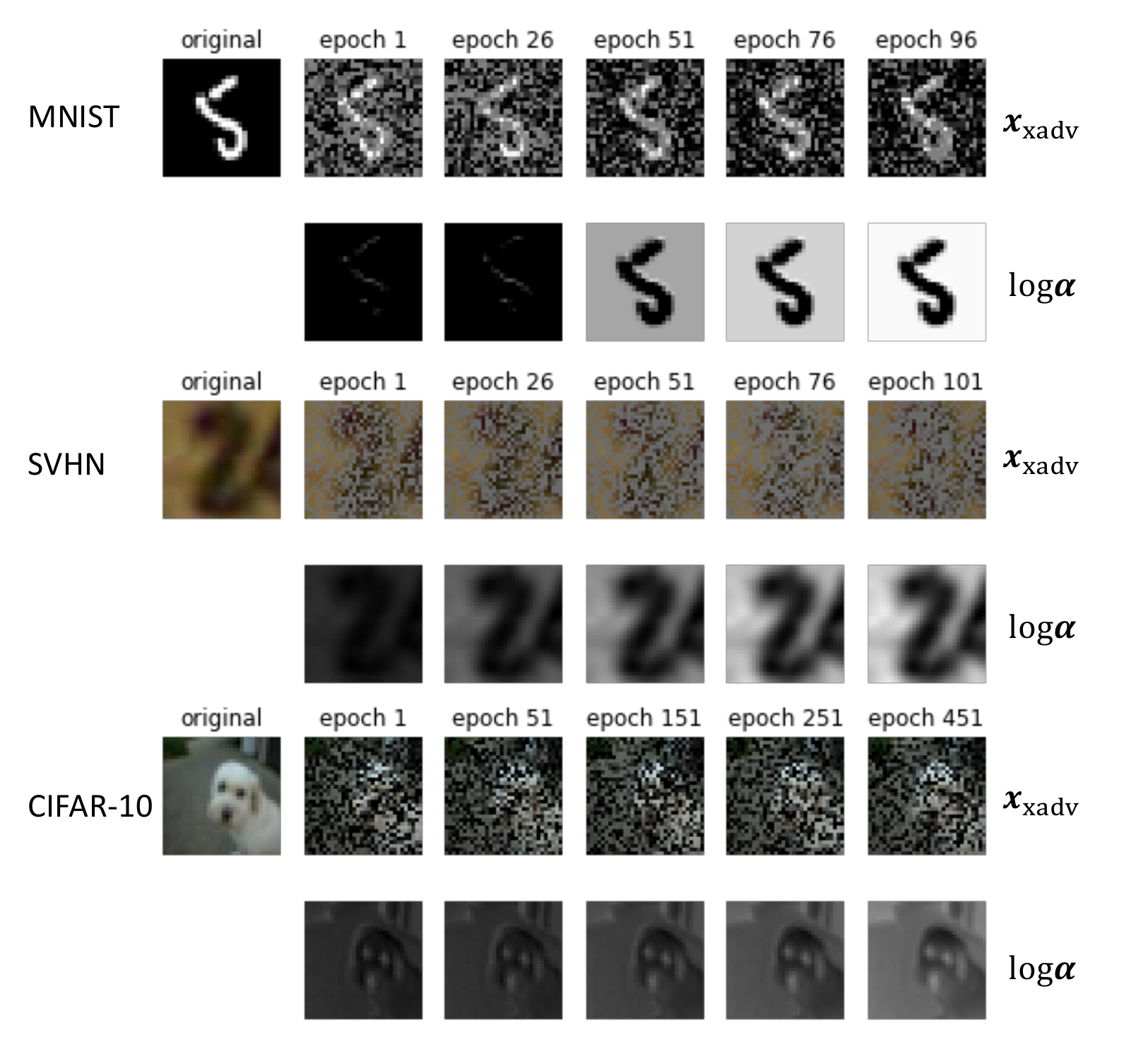}
        \label{process_mask}
    }
    \subfigure[]{
        \includegraphics[width=0.88\columnwidth]{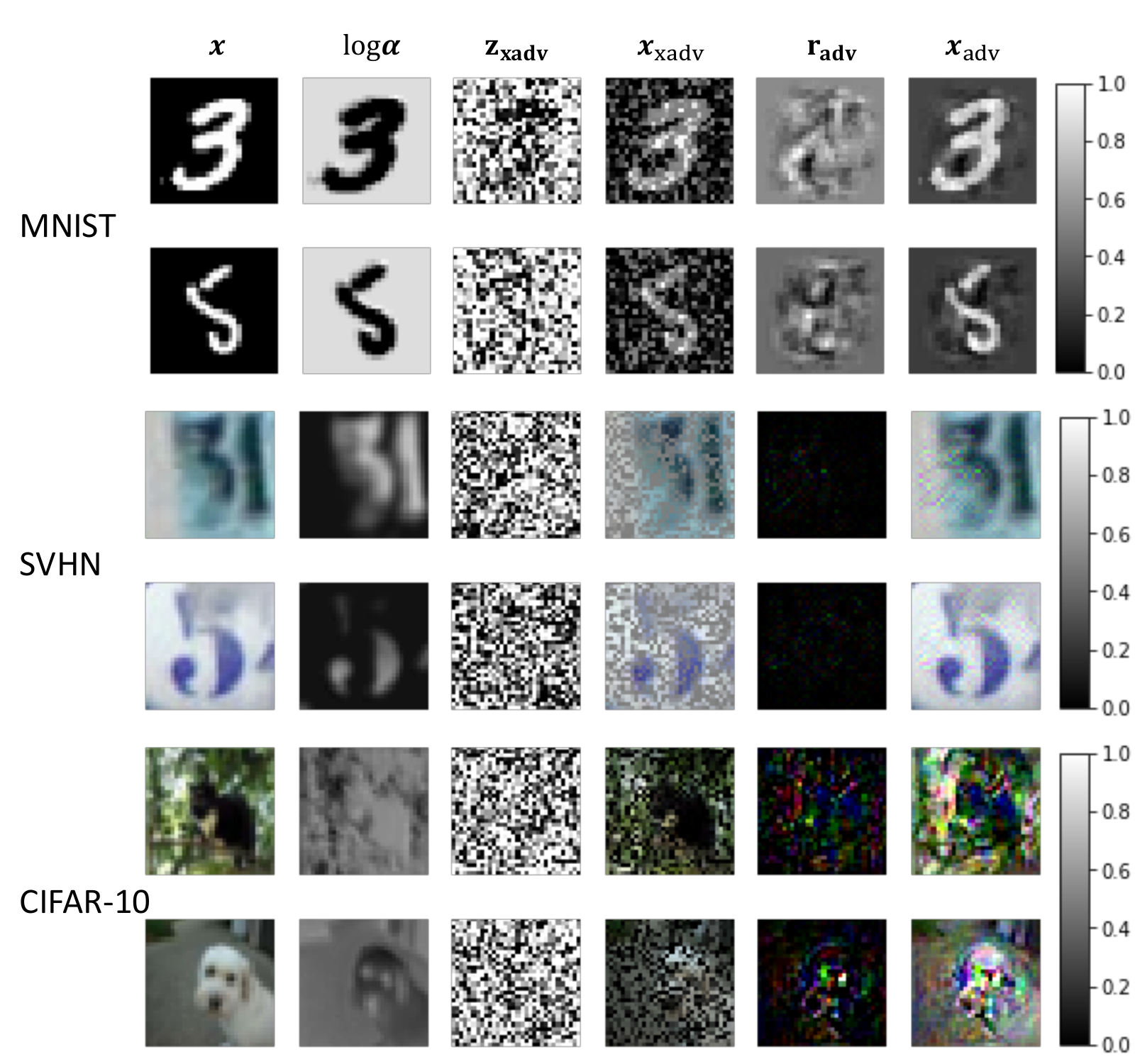}
        \label{vis_adv}
    }
    \vspace{-5pt}
    \caption{Visualization of multiplicative perturbations and additive perturbations from xVAT and VAT. (a) The evolution of $\log\bs{\alpha}$ and $\bs{x}_\textrm{xadv}$ during the training of xVAT on benchmark datasets. (b) Comparison of multiplicative and additive perturbations on example images from benchmark datasets.}
    \label{figure:visualization}
    \vspace{-10pt}
\end{figure*}

\begin{figure*}[ht!]
    \centering
    \subfigure[First convolutional layer]{
        \includegraphics[width=0.58\columnwidth]{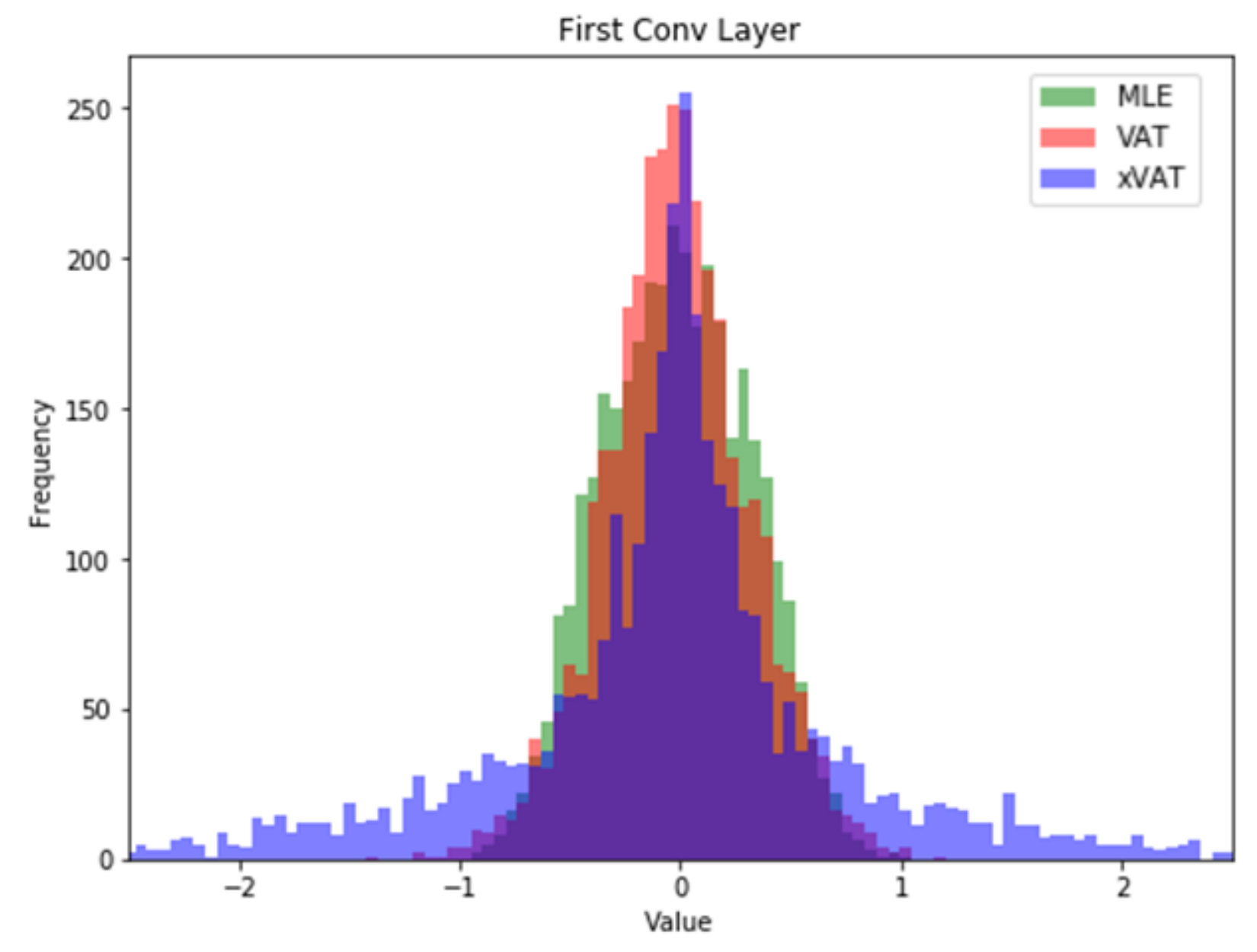}
        \label{figure:first_conv_layer}
    }
    \subfigure[Last convolutional layer]{
        \includegraphics[width=0.58\columnwidth]{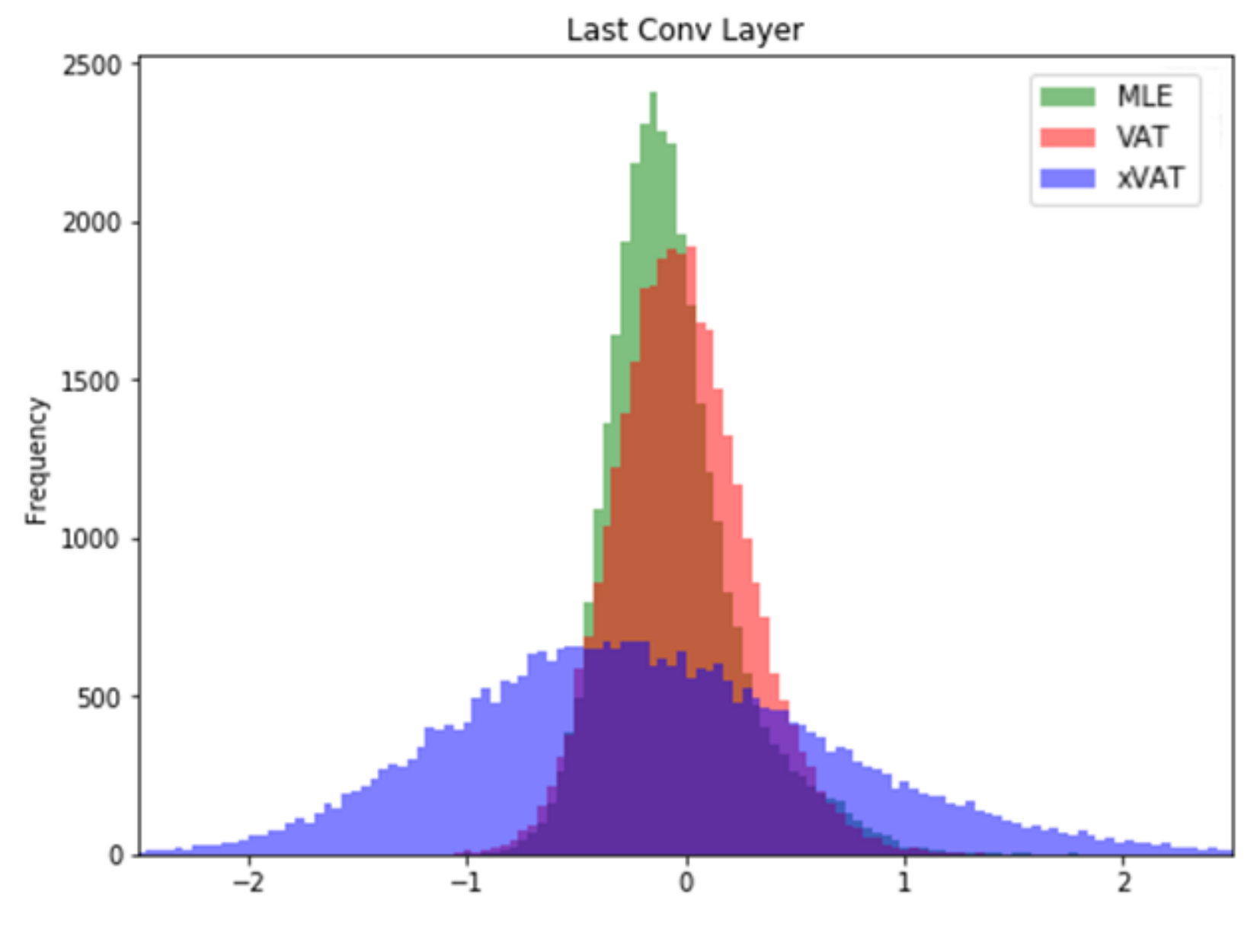}
        \label{figure:last_conv_layer}
    }
    \subfigure[Last fully connected layer]{
        \includegraphics[width=0.58\columnwidth]{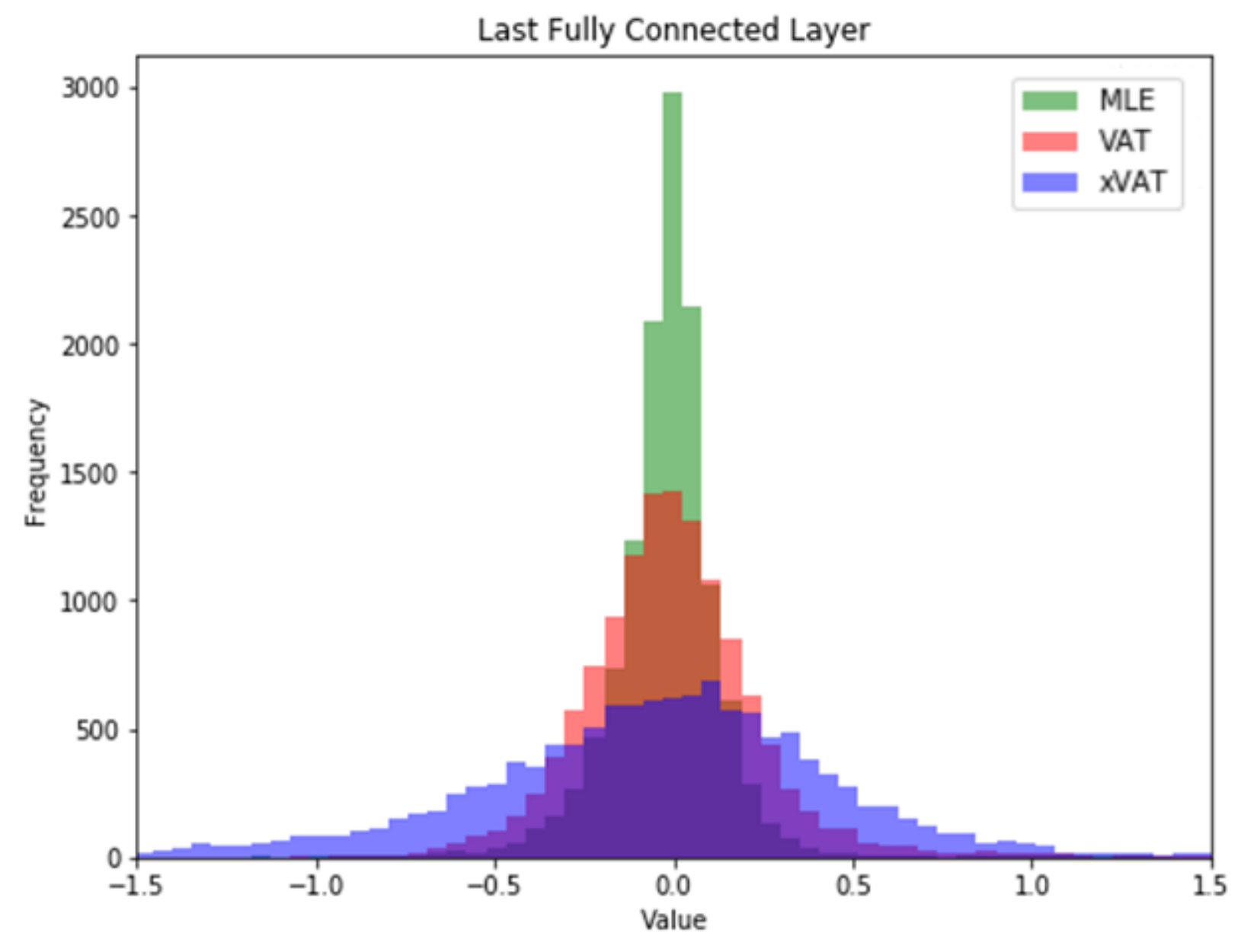}
        \label{figure:linear_layer}
    }
    \vspace{-1mm}
    \caption{Histograms of the classifier weights learned by MLE, VAT and xVAT on CIFAR-100. The histograms are computed from different CNN layers.}
    \label{figure:effect_weight}
    \vspace{-10pt}
\end{figure*}

\subsection{Visualization of Multiplicative Perturbations}
To understand the effectiveness of xVAT in generating multiplicative perturbations, we visualize the evolution of $\log\bs{\alpha}$ and $\bs{x}_\textrm{xadv}$ on example images from MNIST, SVHN and CIFAR-10, with the results shown in Fig.~\ref{process_mask}. As we can see, at the beginning of the xVAT training, the output of sparse mask generator $\log\bs{\alpha}$ isn't very informative as it generates multiplicative perturbations almost uniformly over the entire images. As the training proceeds and the sparse mask generator is trained adversarially, the mask learns to block the salient regions of the input images and leads to more effective adversarial examples. This can be observed from the learned $\log\bs{\alpha}$, which gets more interpretable over training epochs. 

To compare the multiplicative perturbations with the additive perturbations, Fig.~\ref{vis_adv} illustrates their effects on example images from MNIST, SVHN and CIFAR-10. As we can see, visually these two types of perturbations are very different: the multiplicative perturbations are more perceptible than the additive counterparts, but the former are more interpretable than the latter as learned $\log\bs{\alpha}$s clearly discover the salient regions of input images. It is worth mentioning that the more perceptible multiplicative perturbation is not a disadvantage of xAT/xVAT since our work is not to propose new adversarial attack algorithms but to investigate new adversarial training techniques that can improve the robustness of DNNs. Evidently, these results demonstrate that multiplicative perturbations are \emph{at least} as effective as additive perturbations at identifying the blind-spots of DNNs and training on these perturbations can lead more robust DNNs.

%

\subsection{Visualization of Weight Distributions}
To investigate the differences of trained classifiers from different algorithms, we compare the weights of models learned by MLE, VAT and xVAT on CIFAR-100. Fig.~\ref{figure:effect_weight} shows 
the histograms of model weights from three different CNN layers. It is demonstrated that xVAT learns a classifier whose weights have a much higher variance than those learned by VAT and MLE. In other words, xVAT learns a denser classifier from multiplicative perturbations with more non-zero weights than VAT and MLE. As we have demonstrated in Fig.~\ref{figure:eps_in_pert}, the multiplicative perturbations reside inside a rectangle around an input $\bs{x}$, while the additive perturbations can only lie on the ring surrounding $\bs{x}$. As a result, there are more varieties among multiplicative perturbations than their additive counterparts, and therefore the adversarially trained DNNs need more capacities (active neurons) to against multiplicative perturbations.

\section{Conclusion}
In this paper we propose a new type of perturbations that is multiplicative. Compared to the additive perturbations exploited by AT and VAT, the multiplicative perturbations are more perceptible and interpretable. We show that these multiplicative perturbations can be optimized via an $L_0$-norm regularized objective and implemented transductively and inductively. Thanks to the hard concrete reparameterization, the resulting algorithms xAT and xVAT are computationally more efficient than their additive counterparts. Extensive experiments on synthetic and four established image classification benchmarks demonstrate that xAT and xVAT match or outperform the state-of-the-art algorithms while being about 30\% faster. Visualization of xAT and xVAT demonstrates a stark contract to their additive counterparts.

As for future extensions, we are interested in investigating the interplay between xVAT and $L_0$-norm based network sparsification~\cite{l0-arm}. Both algorithms are developed via the $L_0$ regularizations, while xVAT learns a dense network and $L_0$-ARM~\cite{l0-arm} prunes redundant neurons. The combination of them might lead a network that is both robust and efficient.

\section{Acknowledgment}
The authors would gratefully acknowledge the support of VMware Inc. for its university research fund to this research.

\bibliographystyle{IEEEtran}
\bibliography{ml}

\section{Appendix}

%

\subsection{Image Classification Benchmarks}\label{app:datasets}
The image classification benchmarks used in our experiments are described below:
\begin{enumerate}
\item MNIST~\cite{mnist} is a gray-scale image dataset containing 60,000 training images and 10,000 test images of the size $28 \times 28$ for 10 handwritten digits classification.
\item SVHN~\cite{svhn11} is a street view house number dataset containing 73,257 training images and 26,032 test images classified into 10 classes representing digits. Each image may contain multiple real-world house number digits, and the task is to classify the center-most digit. These are RGB Images of the size $32\times 32$.
\item CIFAR-10~\cite{cifar09} contains 10 classes of RGB images of the size $32\times 32$, in which 50,000 images are for training and 10,000 images are for test.
\item CIFAR-100~\cite{cifar09} also has 60,000 RGB images of the size $32\times 32$, except that it contains 100 classes with 500 training images and 100 test images per class.
\end{enumerate}

\subsection{Experimental Setup}\label{app:setup}
Our xAT and xVAT contains two sets of model parameters: (i) classifier parameter $\bs{\theta}$, and (ii) per-image $\log\bs{\alpha}$ parameter as in the transductive implementation or generator parameter $\bs{\theta}_g$ as in the inductive implementation. For the transductive implementation, we optimize $\log \bs{\alpha}$ by SGD with a learning rate of 0.001. For the inductive implementation, we found that one-layer CNN with one $3\times 3$ filter works well for all our experiments, and we optimize the generator by using Adam~\cite{adam2014} with a fixed learning rate of $1e-6$ or $2e-6$. 

As for classifier parameter $\bs{\theta}$, different network architectures are used for different image classification benchmarks. In all our experiments, we use $\lambda = 1, \eta=1$ for xVAT, and $\lambda = 1, \eta=0.5$ for xAT. 

\paragraph{MNIST}
We use an MLP with four hidden layers of 1200, 600, 300 and 150 units, respectively, for the MNIST experiments. The same MLP architecture is used for supervised training and semi-supervised training as in VAT~\cite{vat16,vat18}. The input images are linearly normalized to the range of $[-0.5, 0.5]$. We use the Adam optimizer~\cite{adam2014} with an initial learning rate of 0.002, which is decayed by 0.9 at every 500 iterations. We train the classifier for 50,000 iterations, and at each iteration a mini-batch of 100 labeled images and 250 unlabeled images is used for training. 

\paragraph{CNN Architecture for SVHN, CIFAR-10/100}
On these three benchmarks, for the purpose of fair comparisons, we use the CNN architecture shown in Table~\ref{table:cnn9}, which is the same as the one used in~\cite{vat18,advdrop18,temporalensembling2017}. 

\begin{table}[ht!]
\caption{The CNN architecture used on SVHN, CIFAR-10, and CIFAR-100.}\vspace{-15pt}
\begin{center}
\begin{threeparttable}
\begin{tabular}{ll}
\toprule
Name   & Description                                                      \\
\midrule
Input  & Image                                                            \\
Conv1a & 128 filters, 3 $\times$ 3, Pad=`same', LReLU($\alpha$ = 0.1) \\
Conv1b & 128 filters, 3 $\times$ 3, Pad=`same', LReLU($\alpha$ = 0.1) \\
Conv1c & 128 filters, 3 $\times$ 3, Pad=`same', LReLU($\alpha$ = 0.1) \\
Pool1  & Maxpool 2 $\times$ 2 pixels, Stride 2                        \\
Drop1  & Dropout, p = 0.5                                             \\
Conv2a & 256 filters, 3 $\times$ 3, Pad=`same', LReLU($\alpha$ = 0.1) \\
Conv2b & 256 filters, 3 $\times$ 3, Pad=`same', LReLU($\alpha$ = 0.1) \\
Conv2c & 256 filters, 3 $\times$ 3, Pad=`same', LReLU($\alpha$ = 0.1) \\
Pool2  & Maxpool 2 $\times$ 2 pixels, Stride 2                        \\
Drop2  & Dropout, p = 0.5                                             \\
Conv3a & 512 filters, 3 $\times$ 3, Pad=`valid', LReLU($\alpha$ = 0.1) \\
Conv3b & 256 filters, 1 $\times$ 1, Pad=`same', LReLU($\alpha$ = 0.1) \\
Conv3c & 128 filters, 1 $\times$ 1, Pad=`same', LReLU($\alpha$ = 0.1) \\
Pool3  & Global average pool, 6 $\times$ 6 $\to$ 1 $\times$ 1             \\
Dense  & Fully connected 128 $\to$10                                      \\
Output & Softmax                                                          \\
\bottomrule
\end{tabular}
\end{threeparttable}
\label{table:cnn9}
\end{center}
\vspace{-2mm}
\end{table}

We follow the same normalization schemes of VAT on these three benchmarks. We normalize the SVHN images to the range of $[0, 1]$. For CIFAR-10 and CIFAR-100, we normalize the images with ZCA, based on training set statistics, the same as in VAT~\cite{vat16,vat18}.

In semi-supervised learning, we train the classifier for 50000 iterations on SVHN, 200000 iterations on CIFAR-10, and 120000 iterations on CIFAR-100. At each iteration, we sample a mini-batch of 32 labeled images and 128 unlabeled images. We use the Adam optimizer with an initial learning rate of 0.001 and decay the learning rate linearly at the $[35000, 180000, 80000]$-th iterations. 
 
For supervised learning on CIFAR-10 and CIFAR-100, we train the CNN classifier for 300 epochs with a batch size of 100. We again use Adam with an initial learning rate of 0.003 for xAT and 0.001 for xVAT. We linearly decay the learning rate at the 2nd half of training. 

\subsection{Hyperparameter Tuning}\label{app:hyperparameter}
\begin{figure}[ht]
    \centering
    \includegraphics[width=1.0\columnwidth]{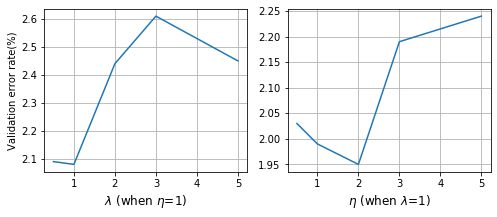}
    \caption{Hyperparameters tuning of $\lambda$ and $\eta$ on the MNIST dataset.}
    \label{figure:hyper}
\end{figure}

Fig.~\ref{figure:hyper} shows the results of hyperparameter tuning of $\lambda$ and $\eta$ on the MNIST dataset. As we can see, $\lambda=1$ and $\eta=1$ yields very competitive classification accuracies. Similar results are observed on the other benchmarks. We therefore use $\lambda=1$ and $\eta=1$ in all our semi-supervised learning tasks.

\end{document}